\documentclass[sigconf, nonacm]{acmart}

\usepackage{xcolor}

\AtBeginDocument{%
  }

\copyrightyear{2025}
\acmYear{2025}
\setcopyright{cc}
\setcctype{by}
\acmConference[CompEd 2025]{Proceedings of the ACM Global Computing Education Conference 2025 Vol 1}{October 21--25, 2025}{Gaborone, Botswana}
\acmBooktitle{Proceedings of the ACM Global Computing Education Conference 2025 Vol 1 (CompEd 2025), October 21--25, 2025, Gaborone, Botswana}
\acmDOI{10.1145/3736181.3747156}
\acmISBN{979-8-4007-1929-5/2025/10}

\begin{document}

\title{Using Sentiment Analysis to Investigate Peer Feedback by Native and Non-Native English Speakers}

\author{Brittney Exline}
\affiliation{%
  \institution{Georgia Institute of Technology}
  \city{Atlanta}
  \state{Georgia}
  \country{USA}}
\email{bexline3@gatech.com}

\author{Melanie Duffin}
\affiliation{%
  \institution{Georgia Institute of Technology}
  \city{Atlanta}
  \state{Georgia}
  \country{USA}}
\email{mduffin@gatech.edu}

\author{Brittany Harbison}
\affiliation{%
  \institution{Georgia Institute of Technology}
  \city{Atlanta}
  \state{Georgia}
  \country{USA}}
\email{bharbison3@gatech.edu}

\author{Chrissa da Gomez}
\affiliation{%
 \institution{Georgia Institute of Technology}
 \city{Atlanta}
 \state{Georgia}
 \country{USA}}
\email{cgomez74@gatech.edu}

\author{David Joyner}
\affiliation{%
 \institution{Georgia Institute of Technology}
 \city{Atlanta}
 \state{Georgia}
 \country{USA}}
\email{david.joyner@gatech.edu}

\renewcommand{\shortauthors}{Exline et al.}

\begin{abstract}
    Graduate-level CS programs in the U.S. increasingly enroll international students, with 60.2\% of master’s degrees in 2023 awarded to non-U.S. students. Many of these students take online courses, where peer feedback is used to engage students and improve pedagogy in a scalable manner. Since these courses are conducted in the US, many students study in a language other than their first. This paper examines how native versus non-native English speaker status affects three metrics related to peer feedback experience in online U.S.-based computing courses. Using the Twitter-roBERTa base model, we perform sentiment analysis of peer reviews written by and to a random sample of 500 students. Then, we analyze how their status as native or non-native English speakers relates to these sentiment analysis scores as well as their overall rating of peer feedback. Results show that native English speakers rate feedback less favorably, while non-native speakers express more positive sentiment in reviews they write but receive less positive sentiment in return. When controlling for sex and age range, significant interactions emerge, suggesting that whether a student natively speaks the language of a course plays a modest but complex role in shaping peer feedback experiences.
\end{abstract}



\maketitle

\section{Introduction}
Since the pandemic, the number of students enrolled in graduate computer science education has grown significantly. The first-time enrollment in graduate programs for mathematics and CS has increased 12.3\% between 2017 and 2022 \cite{mckenzie2023}. Part of this growth can be attributed to the availability of lower-cost online programs \cite{duncan2020}, which allow students to study from anywhere, regardless of the institution’s location. Graduate CS programs in the United States have long had a large portion of enrollment from foreign-born students; 78\% of American universities with Computer Science departments have a majority of graduate students from outside the US \cite{nfap2021}. This indicates that many students who study computer science at more advanced levels are doing so in a language other than their first language. 

With the rise of online education has also come the need for scalable and effective pedagogical tools. One such tool is peer feedback, where students provide formative feedback to their peers on work completed for the class. In the context of CS education, this often involves feedback on conceptual understanding, justification on design decisions, or code explanations submitted in assignments. Peer feedback has been shown to improve student engagement and self-regulated learning \cite{huisman2018}, though this impact can be limited depending on how it is implemented in the digital classroom \cite{zhou2019}. As with any type of feedback in an educational setting, students’ individual characteristics influence how they receive, and experience, said feedback \cite{panadero2022}. 

In this paper, we explore how students learning CS in English as their non-native language engage with and experience peer feedback within the context of online graduate education as compared to native English-speaking students. Using sentiment analysis tools and student survey data, we examine the influence of students’ status as native or non-native English speakers on their subjective rating of peer feedback, and the sentiments expressed and received within peer feedback comments. We aim to address the following research questions:
\begin{itemize}
\item \textbf{RQ1:} How do native and non-native English speakers differ in how they perceive the impact of peer feedback in online CS courses?
\item \textbf{RQ2:} How do native and non-native English speakers differ in the sentiment of the peer feedback they author in online CS courses?
\item \textbf{RQ3:} How do native and non-native English speakers differ in the sentiment of the peer feedback they receive in online CS courses?
\end{itemize}

\section{Literature Review}
This research exists at the intersection of three sub-fields within the broader investigation of the pedagogical role of peer feedback: student characteristics (demographics, prior education, etc.), Native vs. Non-native English speakers, and automated sentiment analysis.

\subsection{Student Characteristics and Peer Feedback}
Many feedback models underscore the role of individual student characteristics, including a model proposed by Panadero and Lipnevich based on a comprehensive review of 14 models \cite{panadero2022}. In this model, student characteristics are the central element because they act as mediators for the other aspects of the model, including agents, which account for providers of feedback beyond the instructor. This model predicts that peers as feedback agents influence how students perceive and react to feedback and that differing student characteristics, including demographic factors, will moderate this effect. 

Many studies investigate the moderating effect of specific student characteristics on how they give and receive feedback, though this work is far from exhaustive. A study of undergraduate psychology students examining gender and feedback training found that female assessors in certain cases give more positive feedback compared to male assessors and tend to write more in their feedback \cite{ocampo2023}. Another study, measuring perceived learning after students revised essays based on peer feedback, found that gender did not play a significant role in students’ level of perceived learning, but level of education (graduate vs. undergraduate) did significantly impact perceived learning \cite{noroozi2024}. The interaction of different student characteristics and the context of giving or receiving feedback can also be surprising. In the context of STEM education, a study found that when giving peer feedback anonymously, female students gave significantly more negative feedback as compared to the non-anonymous condition, while there was no such effect for male students \cite{lane2018}. This theory, along with the selection of these studies, shows that there are many ways students’ backgrounds and circumstances can influence their relationship to peer feedback. Exploring how native and non-native speakers differ in their engagement with peer feedback in the unique context of online graduate CS programs taught in English can add to this body of knowledge.

\subsection{Native vs. Non-native English Speakers and Peer Feedback}
Peer feedback has widely been seen as effective in improving students’ writing skills in an ESL context \cite{baharudin2021}, however, less research has been done comparing native and non-native English speakers. There is evidence of different language patterns employed by native vs. non-native English speakers that may impact the sentiment in their written peer feedback. A study comparing peer reviews from three institutions from different countries, all conducted in English, showed that students from the American institution used fewer cognitive verbs (i.e. think, feel) and fewer hedges in their reviews compared to non-American institutions, indicating differences in how non-native speakers negotiate criticism and praise \cite{warnsby2018}. There is also evidence for differences in how peer feedback impacts the learning of native and non-native speakers; in a study of student teachers' attitudes toward peer feedback, non-native speakers were less inclined to trust peer feedback before engaging in it, but afterward, non-native speakers were more likely to rate peer feedback as helpful in improving their work \cite{mcgarrell2010}. Non-native speakers have been found to have different patterns in the written feedback they provide compared to native speakers; non-native speakers gave less evaluative and corrective feedback in mixed peer feedback groups containing native and non-native speakers \cite{zhu2001}.

\subsection{Sentiment Analysis and Peer Feedback}
Peer feedback text can be analyzed in many ways, including through manual coding or through machine learning-based tools. Coding schemes used for manual coding often emphasize some aspects of sentiment; for example, verification refers to statements about the correctness of the work, and these tend to be expressed positively, negatively, or neutrally \cite{gielen2015}. ML techniques can also be used to extract sentiment from text, and this has been applied to the study of peer feedback and the implications for student experiences of peer feedback in an online setting \cite{beasley2021, huang2023, lane2018}. One study used sentiment analysis to better understand student engagement in data visualization courses, finding that students with higher performance in the class tended to have more positive sentiment words in their peer feedback than those with lower performance, and that seniors and graduate students tended to have fewer sentiment words per review, indicating lower engagement \cite{beasley2021}. In a study of Chinese education students, the researchers developed their own model to analyze peer feedback, building off other language models including BERT, which they found to be relatively accurate for their uses \cite{huang2023}. They tied the results back to students’ experiences of online education burnout, finding that receiving negative feedback correlated with higher burnout \cite{huang2023}. 

\subsection{Study Context}
This study takes place in the context of a large, affordable, online graduate program in computer science. There are a number of features of this program and its student body key to understanding this paper's findings. First, the program is relatively affordable (by U.S. standards) at under 7,000 USD for the entire degree, drawing in a more varied student body: many students enroll for personal edification rather than career benefits given the low cost, and students may enroll with a lower advanced commitment to complete the program given that dropping out does not yield massive student debt. Second, the program is roughly 32\% non-U.S. residents and roughly 33\% non-U.S. born citizens, representing a significant international contingent. Third, the program's admissions are highly inclusive, accepting any student meeting minimum requirements with no capacity constraints; therefore, the range of backgrounds and aptitudes among the student body is higher than highly selective in-person programs.

Additionally, the nature of the peer review tool and accompanying activity is relevant to this study. The tool is a semi-proprietary tool originally developed at the university before it spun off into a separate entity. Peer-assigned scores do \emph{not} affect the recipients' actual class grades; peer review is solely for feedback and learning. Feedback is thus non-anonymous as its intention is more as a community-forming exercise than a reliable score-assignment exercise. Students are graded on the quality of the reviews that they \emph{write} to ensure substantive engagement, but the scores and feedback students \emph{receive} have no direct impact on their scores.

\section{Methodology}
Our methodology consisted of two primary phases: first, drawing and cleaning a dataset of student demographics in the course, and second, calculating a positivity score for the peer reviews each student received during the course.

\subsection{Dataset}
This data was collected from two sources for past semesters of three classes in an online graduate program in Computer Science spanning Fall 2018 to Summer 2022, with approval from the institution’s IRB which covers all the collected data described below. From the program's learning management system, we obtained student-level information such as their final grade and voluntary self-reported survey responses collected throughout each semester. From the program's homegrown peer review system, we collected the comments students wrote each other as feedback on written assignments they submitted each week during each class. These data were merged on student email addresses that were then replaced with anonymized identifiers for subsequent analysis. Students who did not submit the surveys necessary to perform the analysis were removed from the dataset, resulting in 7,101 total students. From this set, we randomly selected 500 students for analysis, along with the ~41,000 feedback comments that these students wrote and received from other students. We chose to sample from the population to balance having a large body of feedback comments for the sentiment analysis tool to analyze without the cost of running it on the full set of over 500,000 comments.

Analysis was constrained to the values gathered by the university as part of administering the class. The surveys offered four options for gender (Female, Male, Other, Prefer not to say), but in practice over 99\% of students selected either Male or Female, leaving insufficient data to draw any analysis on those that selected another option. For age range, course surveys only offered age ranges rather than specific numbers. For English fluency, course surveys offered four options (Native speaker, Fully fluent non-native speaker, Partially fluent, and non-fluent), but again in practice over 98\% of students selected either Native speaker or Fully fluent non-native speaker, leaving insufficient data to specifically analyze Partially fluent and non-fluent speakers.

Based on these constraints, the items available for analysis were:
\begin{itemize}
\item Sex: Male, Female
\item Age Range: 18--24, 25--34, 35--44, 45--54, 55+
\item English Fluency: Native speaker, Non-native speaker
\item Likert-scale rating of the peer feedback component of the course (On a scale of 1 to 7, please rate your agreement with the statement: "The Peer Feedback system has improved my experience in this class."), asked at the end of the course. This item is open to interpretation by the students, inclusive of their perception of the feedback received by other peers, the activity of giving peer feedback, and the ease of use of the system, as well as their enjoyment of these aspects or how valuable they perceived it to be for their learning.
\end{itemize}

We recognize that these discrete labels do not capture the spectra of gender and English fluency; our goal here is to identify trends within the confines of the data we \emph{do} have to inform subsequent data that may investigate these parameters with greater nuance.

\subsection{Sentiment Analysis of Peer Feedback}
\begin{figure}[ht]
    \centering
    \includegraphics[width=1\linewidth]{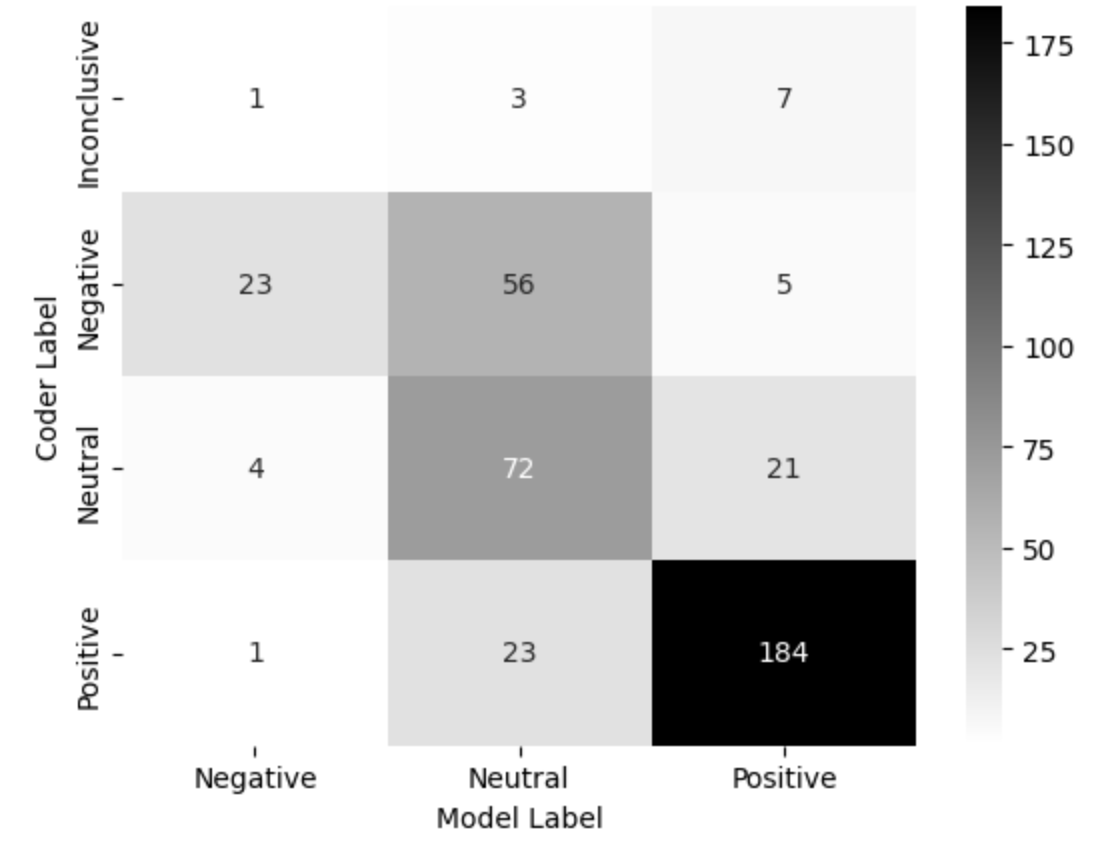}
    \caption{Confusion Matrix of Labels from Human Coders vs. The Model}
    \label{confusion matrix}
    \Description{A confusion matrix showing the agreement between labels from human coders and predictions made by the model.}
\end{figure}

In addition to the survey items, our analysis calculated the relative positive sentiment of the feedback written and received by the students. We decided to use sentiment analysis to process the large volume of peer feedback data. 
We evaluated various sentiment analysis models used for similar use cases to determine the sentiment of single sentences in prior work \cite{koonchanok2024, neumann2021, newman2018} and selected twitter-roBERTa-base. Although VADER showed competitive results \cite{koonchanok2024}, it performed worse on our data, and we favored roBERTa-base over other options as it is more recognized as a baseline for English-language sentiment analysis. To better determine the accuracy for our peer feedback comment data, 400 random sentences were selected and labeled as positive, neutral, or negative independently by each of four researchers on the team. We labeled sentences as positive that contained evaluative feedback that praised the work (i.e. “Good job!”, “I like how you...”) or contained language meant to convey politeness (i.e. “Thank you for sharing.”). Neutral and negative sentiments were assigned similarly based on the evaluations of the work. These categorizations align with the verification feedback style defined by Gielen and DeWever \cite{gielen2015}. Because sentences could sometimes contain separate phrases with differing sentiments, we marked sentences with a positive and a negative phrase as neutral, a positive and a neutral phrase as positive, and a neutral and negative phrase as negative. The sentiment analysis model also categorized the sentences, and these labels were compared across the four human coders and the model. All four human coders decided on the same label for 65.5\% of the sentences, and 3 of 4 agreed on the same label for 97.25\% of the sentences, leaving 2.75\% of sentences as inconclusive. The sentiment analysis matched the human coders’ label 69.75\% of the time when the inconclusive sentences were considered mismatches, and 72.5\% of the time if they were considered matches. Figure \ref{confusion matrix} below details the breakdown of matches and mismatches across labels.

Because peer feedback reviews can range in size and can contain varying sentiments throughout the writing, the sentiment analysis tool was run on each sentence of each review. Each review was given a single sentiment score by averaging the sentiment of the sentences in the review; -1 was given for negative sentences, 0 for neutral, and 1 for positive sentences. For example, if a 4-sentence review had two positive, one neutral, and one negative sentence, its score would be .25 ((1+1-1)/4). Then, two scores were generated for each student in the data set; the first score averaged the sentiment scores of reviews \emph{written by} the student, and the second averaged the sentiment scores of reviews \emph{received by} the student from others. If the student either did not write any reviews or did not receive any reviews, they were dropped from the data set, resulting in 474 total students included in the analysis.

\subsection{Data Analysis}
First, the Kruskal-Wallis test is performed to detect statistically significant differences between levels of English fluency for each of the dependent variables: the final peer feedback rating, the relative sentiment score of peer feedback by the student, and the relative sentiment score of peer feedback to the student. Next, to see if there are still significant differences when controlling for gender and age range, a 3-way ANOVA is performed for each dependent variable.

\section{Results}
\subsection{RQ1: Native vs. Non-native English Speakers and Peer Feedback Rating}
Native English-speaking students rated peer feedback less favorably (mean = 4.46) than non-native English-speaking students who rated it 10\% higher on average (mean = 4.91), and this difference was statistically significant (H = 8.41, p = 0.004, g = 0.319) (see Table~\ref{tab:rating-flu-native}). After controlling for the main effects of gender and age range, there was still a statistically significant difference between native and non-native English speakers (F = 4.20, p = 0.041). There was also a significant difference between age groups, where ratings became higher as the age range increased. (F = 2.84, p = 0.024).

\begin{table}[ht]
  \caption{Differences in Peer Feedback Rating (from 1 to 7) between Native and non-native speakers}
  \begin{tabular}{lp{1cm}p{1.5cm}p{1.5cm}}
    \toprule
    English Fluency&N&Mean&SD\\
    \midrule
    Native Speakers& 287& 4.46&1.41\\
    Non-native Speakers& 187& 4.91&1.41\\
    Total& 474& 4.64&1.41\\
    \bottomrule
\end{tabular}
  \label{tab:rating-flu-native}
\end{table}

\begin{figure}[ht]
    \centering
    \includegraphics[width=1\linewidth]{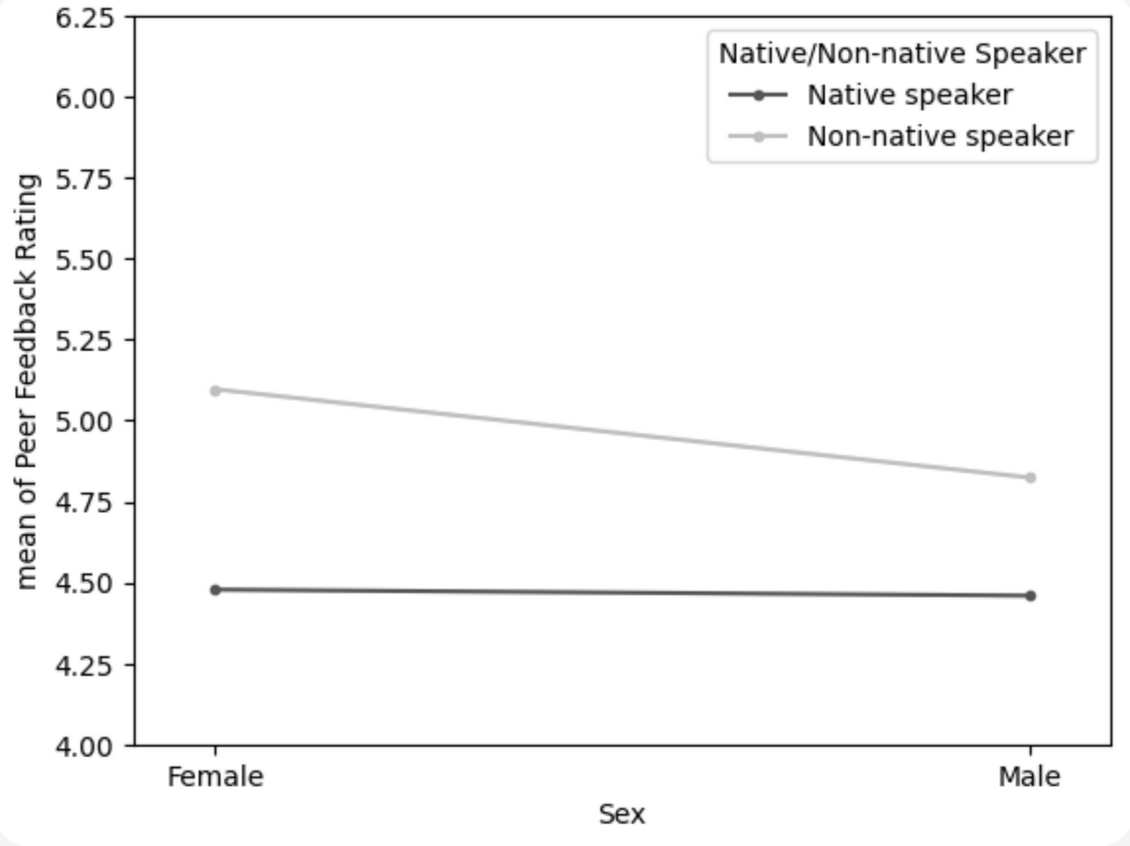}
    \caption{Interaction Plot for Peer Feedback Rating: Native/Non-native Speakers and Sex}
    \label{rating-flu-gender}
    \Description{An interaction plot showing peer feedback ratings as a function of Native vs. Non-native English speakers and sex.}
\end{figure}

\begin{figure}[ht]
    \centering
    \includegraphics[width=1\linewidth]{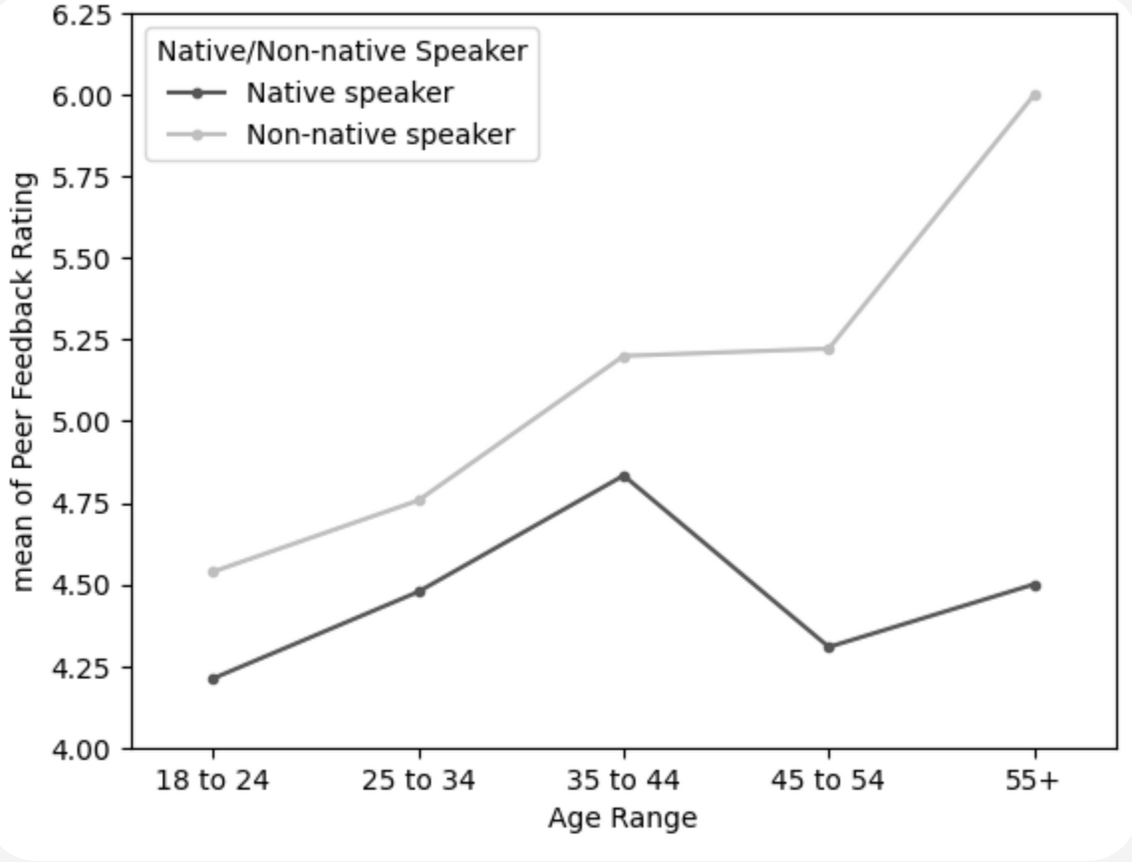}
    \caption{Interaction Plot for Peer Feedback Rating: Native/Non-native Speakers and Age Range }
    \label{rating-flu-age}
    \Description{An interaction plot showing peer feedback ratings as a function of Native vs. Non-native English Speakers and age range.}
\end{figure}

\subsection{RQ2: Native vs. Non-native English Speakers and Relative Sentiment Expressed by Students}
Non-native English-speaking students had significantly higher relative positive sentiment scores, higher by 13\% on average (mean = 0.465), in the peer feedback that they authored as compared to native English speakers (mean = 0.413, H = 8.71, p = 0.003, g = 0.285) (see Table~\ref{tab:rel-sent-fl-exp}). This effect holds after controlling for the main effects of gender and age range (F = 8.98, p = 0.003). Sex also showed significant differences in sentiment expressed, where female students had a higher score than male students (F = 5.66, p = 0.018).

\begin{table}[ht]
\centering
  \caption{Differences in Relative Sentiment Expressed (from -1 to 1) between Native and non-native speakers}
  \begin{tabular}{lp{1cm}p{1.5cm}p{1.5cm}}
    \toprule
    English Fluency&N&Mean&SD\\
    \midrule
    Native Speakers& 287& 0.413&0.179\\
    Non-native Speakers& 187& 0.465&0.188\\
    Total& 474& 0.433&0.184\\
    \bottomrule
\end{tabular}
  \label{tab:rel-sent-fl-exp}
\end{table}

\begin{figure}
    \centering
    \includegraphics[width=1\linewidth]{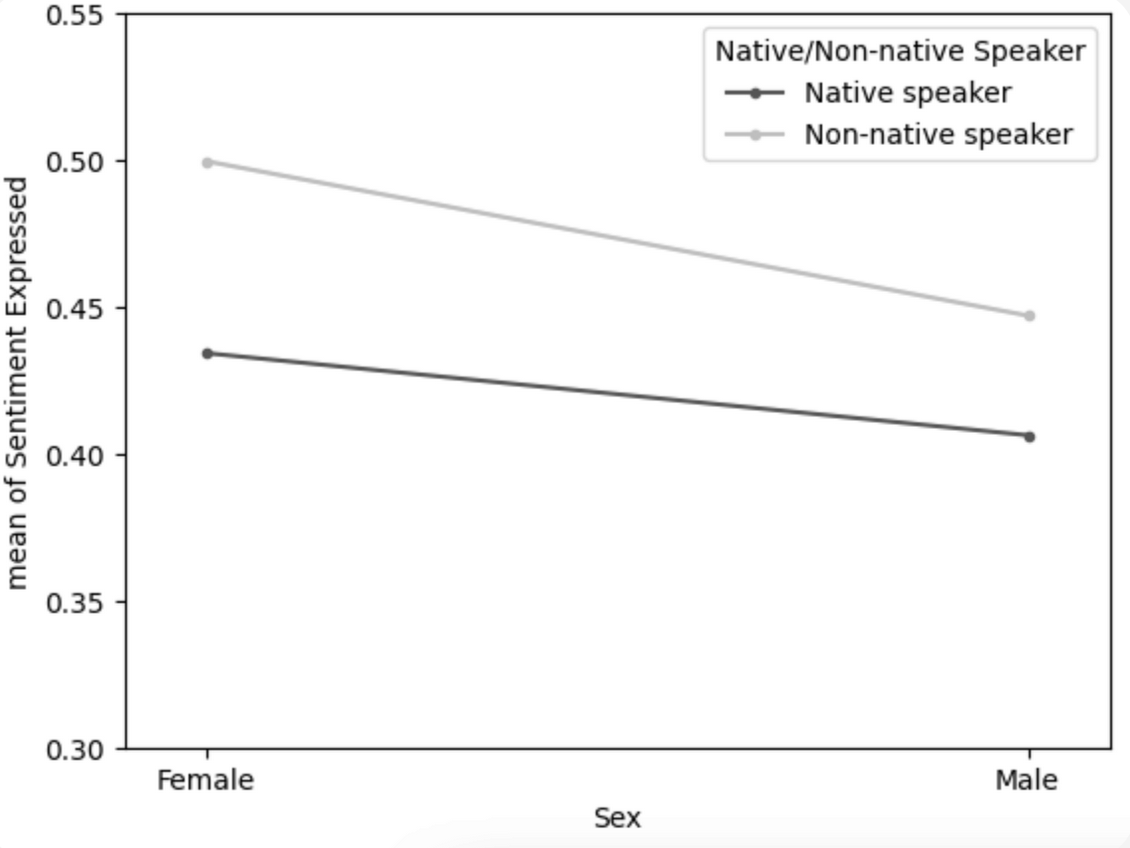}
    \caption{Interaction Plot for Relative Sentiment Expressed: Native/Non-native Speakers and Sex}
    \label{sent-exp-flu-gender}
    \Description{An interaction plot showing sentiment expressed as a function of Native vs. Non-native English speakers and sex.}
\end{figure}

\begin{figure}
    \centering
    \includegraphics[width=1\linewidth]{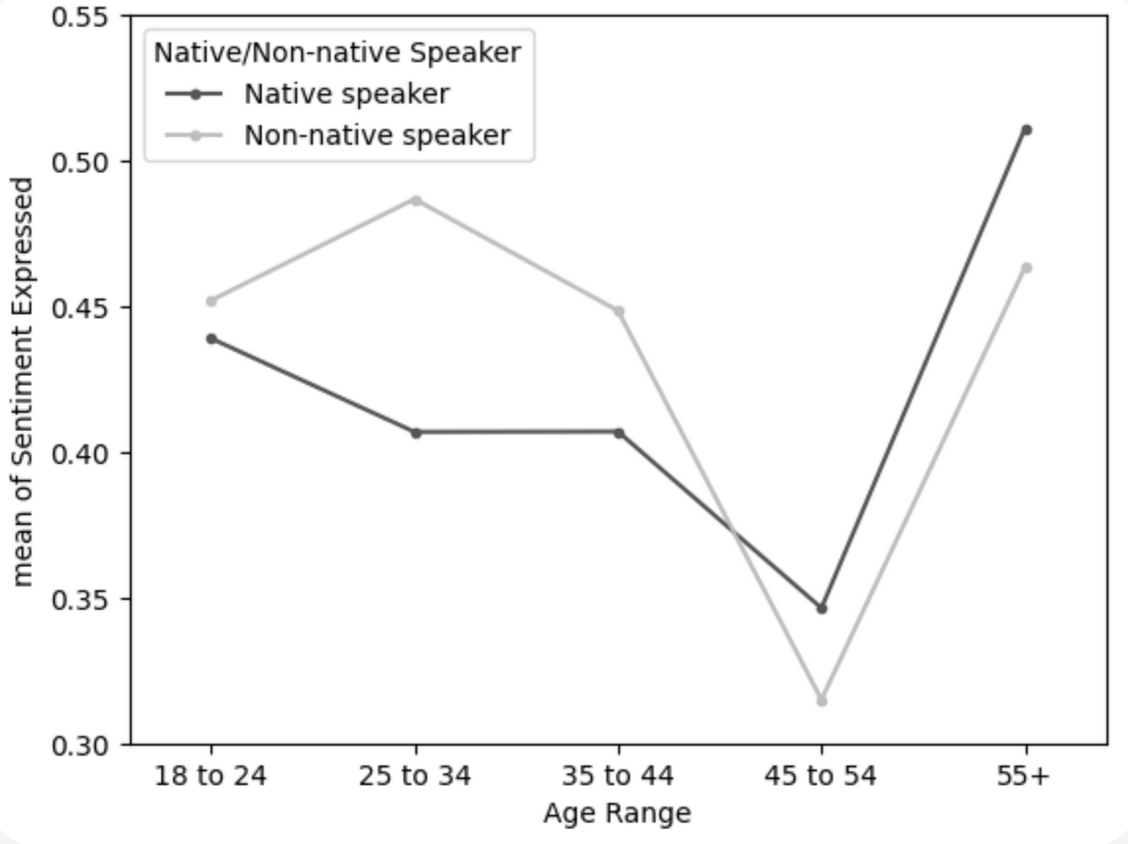}
    \caption{Interaction Plot for Relative Sentiment Expressed: Native/Non-native Speakers and Age Range}
    \label{sent-exp-flu-age}
    \Description{An interaction plot showing sentiment expressed as a function of Native/Non-native English speakers and age range.}
\end{figure}

\subsection{RQ3: Native vs. Non-native English Speakers and Relative Sentiment Received by Students}
Native English-speaking students received significantly higher relative positive sentiment scores, higher by 4\% on average (mean = 0.461), in their peer feedback as compared to non-native English speakers (mean = 0.441, H = 4.82, p = 0.028, g = 0.22) (see Table~\ref{tab:rel-sent-fl-rec}). After controlling for the main effects of gender and age range, there remains a significant difference (F = 5.03, p = 0.025). There are also statistically significant interaction effects with native vs. non-native English speakers, sex, and age range (F = 2.74, p = 0.028).

\begin{table}[ht]
  \caption{Differences in Relative Sentiment Received (from -1 to 1) between Native and non-native speakers}
  \begin{tabular}{lp{1cm}p{1.5cm}p{1.5cm}}
    \toprule
    English Fluency&N&Mean&SD\\
    \midrule
    Native Speakers& 287& 0.461&0.085\\
    Non-native Speakers& 187& 0.441&0.099\\
    Total& 474& 0.453&0.092\\
    \bottomrule
\end{tabular}
  \label{tab:rel-sent-fl-rec}
\end{table}

\begin{figure}
    \centering
    \includegraphics[width=1\linewidth]{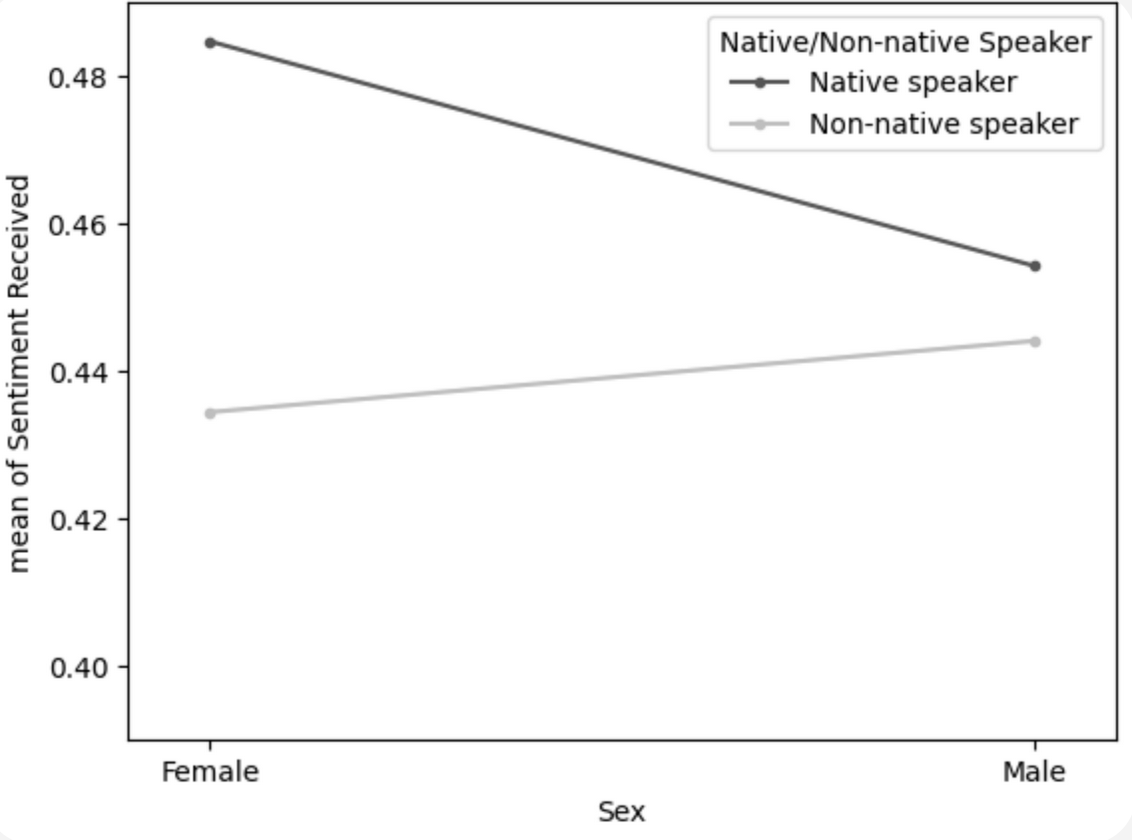}
    \caption{Interaction Plot for Relative Sentiment Received: Native/Non-native Speakers and Sex}
    \label{sent-rec-flu-gender}
    \Description{An interaction plot showing sentiment received as a function of Native vs. Non-native English speakers and sex.}
\end{figure}

\begin{figure}
    \centering
    \includegraphics[width=1\linewidth]{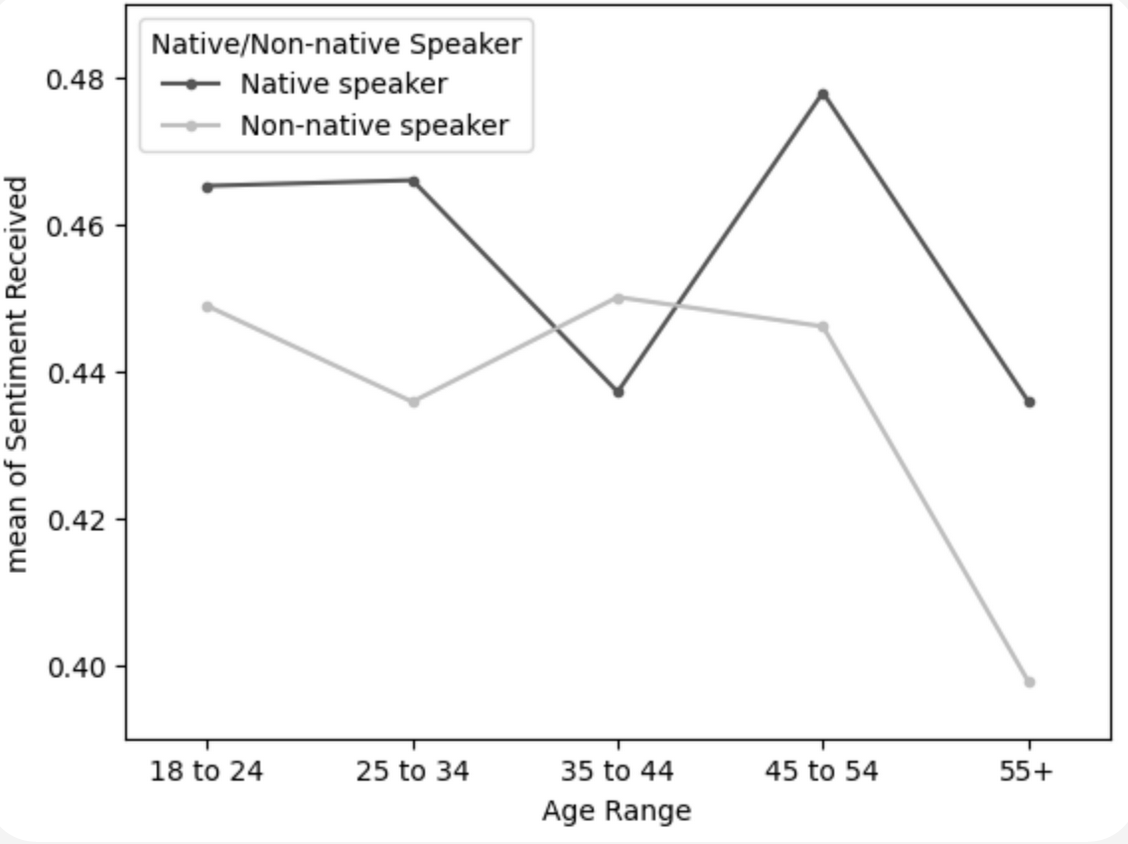}
    \caption{Interaction Plot for Relative Sentiment Received: Native/Non-native Speakers and Age Range}
    \label{sent-rec-flu-age}
    \Description{An interaction plot showing sentiment received as a function of Native vs. Non-native English speakers and age range.}
\end{figure}

\section{Discussion}
Our findings show that native and non-native English speakers have significantly different perceptions and experiences of peer feedback in online graduate CS courses, even if these differences are modest in scale. In addressing RQ1, non-native speakers appear to view peer feedback more favorably in the class than do native English speakers. However, we do not have insight into why this may be the case, though it aligns with some of the prior research presented in our literature review \cite{mcgarrell2010}. A limitation of our data is the reliance on a non-validated survey item that does not distinguish between distinct aspects of peer feedback that students may be responding to, such as the benefits of seeing others’ work, the usefulness of feedback received, or the ease of use of the system. Age also impacted students’ perception of peer feedback, where older non-native English-speaking students tended to find peer feedback most useful. 

Regarding RQ2, we found that non-native English speakers tended to have more positive sentiment in the feedback they wrote compared to English speakers. Since reviews were randomly assigned each week, this is unlikely to be due to differences in the quality of work that students were reviewing. This finding is consistent in part with research into how native and non-native speakers express disagreement; in some cases, non-native speakers may choose to avoid expressing dissenting opinions altogether \cite{kreutel2007}. Gender also played a significant role, with female students, regardless of English fluency, tending to have higher positive sentiment in the peer feedback they wrote compared to male students. This is consistent with some of the studies covered in the literature review \cite{lane2018, ocampo2023}. This is also corroborated by communication studies that suggest men use more assertive language than women, which includes language “disagreeing with or criticizing the other’s contributions” \cite{leaper2007}. The same review also found that women use more affiliative language than men, which includes “showing support, expressing agreement, and acknowledging the other’s contributions” \cite{leaper2007}. These definitions of assertive language and affiliative language correlate with negative and positive sentiment respectively as defined by the human reviewers, which correlate with the sentiment analysis tool’s evaluations. One limitation here is that sentiment is only one aspect of feedback, and so insights into the differences in feedback-giving behavior are constrained. 

Our final result, related to RQ3, was that non-native English speakers tended to receive less positive sentiment when receiving peer feedback compared to native speakers. This difference is modest but could indicate biases in how the writing of non-native speakers is perceived. In a review of studies of US-based faculty responses to work from native and non-native speakers, the author found evidence of linguistic bias \cite{reichelt2021}. Faculty reactions were mixed; several studies found evidence that higher grades were given to the same work thought to be by ESL students compared to native speakers \cite{reichelt2021}. But in one study, faculty gave richer marginal feedback to papers by native speakers than to those by ESL students \cite{rubin1997}. However, differences between faculty reactions and peer reactions to non-native English speakers were pronounced in a study of multilingual ESL immigrant students, who described faculty as supportive but felt marginalized in group settings with White native-English speaking students \cite{nielsen2014}. Complicating this finding is the fact that significant interactions across all three of the demographic variables were found, indicating that even in an online setting, biases related to age, gender, and English fluency (or nationality) may be present in how CS students provide peer feedback.

\section{Limitations}
\balance
As mentioned in the discussion, our survey item assessing students’ perceptions of the peer feedback system is conflating several distinct aspects of peer feedback that students experience. This is a limitation of our archival dataset, where the survey items given to students were primarily designed to capture data relevant to improving the class. Future work could address this by introducing more specific and ideally validated survey items or employ qualitative analysis of open text responses. 

The data was collected between Fall 2018 and Summer 2022, during which the use of generative AI tools (e.g., ChatGPT, GitHub Copilot) in academic settings was either non-existent or very limited. As such, the peer feedback from this time frame was likely authored entirely by students without assistance from AI. With the rapid rise of AI tools in 2023 and beyond, peer feedback has since been shaped by AI-assisted writing, potentially altering the tone, sentiment, and structure of feedback.

Another limitation in this study relates to the calculation of relative overall sentiment of a student’s peer feedback text over the course of a class. Our methodology would give the same score to peer feedback text that contained all neutral sentences as it would to text that contained an equal number of positive and negative sentences. This fails to capture the proportion of positive, neutral, and negative sentiment sentences used by the students. Our sentiment analysis tool has the additional limitation of an accuracy rate around 70\%, which may be due to differences between our dataset and the dataset consisting of millions of tweets that the tool was originally trained on. It may skew the scores more positive as the most common error of the sentiment analysis tool was to classify negative sentences as neutral. Sentiment only captures one aspect of feedback, and either qualitative analysis of a smaller dataset or training of an ML algorithm that can better extract other relevant features of feedback would be needed to better contextualize these sentiment-related findings. 

\section{Conclusion and Future Work}
Student characteristics play an important moderating role in how students perceive and engage with peer review. In this paper, we find that there are significant differences in how non-native English speakers engage with peer feedback. Though they tend to rate peer feedback as more helpful than their native English-speaking peers, they also tend to express more positive sentiment in their feedback while receiving feedback with less positive sentiment. Other demographic characteristics such as sex and age influence this experience in subtle but notable ways. In continuing the pursuit to make CS more inclusive and equitable, the presence of these differences, particularly in an online environment with adult learners, should be accounted for when designing peer feedback activities. Future work could better capture the reasons non-native speakers find peer feedback to positively impact their course experience as compared to native speakers. More sophisticated ML algorithms could be developed to understand other aspects of feedback beyond sentiment, and differences between native and non-native English-speaking students could be further examined. Understanding both the subjective experience of students engaging in peer feedback and the qualities of the peer feedback artifacts that they engage with can inform designing and testing different peer feedback interventions (i.e. peer feedback groups, peer feedback training). Measuring the differences in how students engage with peer feedback in this setting is the first step to understanding how to improve peer feedback, especially as these online CS programs continue to grow and scale.

\bibliographystyle{ACM-Reference-Format}
\bibliography{main-bib}


\begin{thebibliography}{24}


\ifx \showCODEN    \undefined \def \showCODEN     #1{\unskip}     \fi
\ifx \showISBNx    \undefined \def \showISBNx     #1{\unskip}     \fi
\ifx \showISBNxiii \undefined \def \showISBNxiii  #1{\unskip}     \fi
\ifx \showISSN     \undefined \def \showISSN      #1{\unskip}     \fi
\ifx \showLCCN     \undefined \def \showLCCN      #1{\unskip}     \fi
\ifx \shownote     \undefined \def \shownote      #1{#1}          \fi
\ifx \showarticletitle \undefined \def \showarticletitle #1{#1}   \fi
\ifx \showURL      \undefined \def \showURL       {\relax}        \fi
\providecommand\bibfield[2]{#2}
\providecommand\bibinfo[2]{#2}
\providecommand\natexlab[1]{#1}
\providecommand\showeprint[2][]{arXiv:#2}

\bibitem[Baharudin and Razali(2021)]%
        {baharudin2021}
\bibfield{author}{\bibinfo{person}{Muhammad~Danial Baharudin} {and} \bibinfo{person}{Abu~Bakar Razali}.} \bibinfo{year}{2021}\natexlab{}.
\newblock \showarticletitle{A Review of Literature on the Potentials and Problems of Face-to-Face and Online Peer Feedback and the Patterns of Interaction among ESL/EFL Learners in a Peer Feedback Environment}.
\newblock \bibinfo{journal}{\emph{3L The Southeast Asian Journal of English Language Studies}} \bibinfo{volume}{27}, \bibinfo{number}{4} (\bibinfo{year}{2021}), \bibinfo{pages}{114--128}.
\newblock
\href{https://doi.org/10.17576/3l-2021-2704-09}{doi:\nolinkurl{10.17576/3l-2021-2704-09}}


\bibitem[Beasley et~al\mbox{.}(2021)]%
        {beasley2021}
\bibfield{author}{\bibinfo{person}{Zachariah~J. Beasley}, \bibinfo{person}{Alon Friedman}, {and} \bibinfo{person}{Paul Rosen}.} \bibinfo{year}{2021}\natexlab{}.
\newblock \showarticletitle{Through the Looking Glass: Insights Into Visualization Pedagogy Through Sentiment Analysis of Peer Review Text}.
\newblock \bibinfo{journal}{\emph{IEEE Computer Graphics and Applications}} \bibinfo{volume}{41}, \bibinfo{number}{6} (\bibinfo{year}{2021}), \bibinfo{pages}{59--70}.
\newblock
\href{https://doi.org/10.1109/mcg.2021.3115387}{doi:\nolinkurl{10.1109/mcg.2021.3115387}}


\bibitem[Duncan et~al\mbox{.}(2020)]%
        {duncan2020}
\bibfield{author}{\bibinfo{person}{Alex Duncan}, \bibinfo{person}{Bobbie Eicher}, {and} \bibinfo{person}{David~A. Joyner}.} \bibinfo{year}{2020}\natexlab{}.
\newblock \showarticletitle{Enrollment Motivations in an Online Graduate CS Program}. In \bibinfo{booktitle}{\emph{Proceedings of the 51st ACM Technical Symposium on Computer Science Education}}. \bibinfo{publisher}{ACM}, \bibinfo{address}{Portland, OR, USA}, \bibinfo{pages}{1241--1247}.
\newblock
\href{https://doi.org/10.1145/3328778.3366848}{doi:\nolinkurl{10.1145/3328778.3366848}}


\bibitem[Gielen and Wever(2015)]%
        {gielen2015}
\bibfield{author}{\bibinfo{person}{Mario Gielen} {and} \bibinfo{person}{Bram~De Wever}.} \bibinfo{year}{2015}\natexlab{}.
\newblock \showarticletitle{Structuring peer assessment: Comparing the impact of the degree of structure on peer feedback content}.
\newblock \bibinfo{journal}{\emph{Computers in Human Behavior}}  \bibinfo{volume}{52} (\bibinfo{year}{2015}), \bibinfo{pages}{315--325}.
\newblock
\href{https://doi.org/10.1016/j.chb.2015.06.019}{doi:\nolinkurl{10.1016/j.chb.2015.06.019}}


\bibitem[Huang et~al\mbox{.}(2023)]%
        {huang2023}
\bibfield{author}{\bibinfo{person}{Changqin Huang}, \bibinfo{person}{Yaxin Tu}, \bibinfo{person}{Zhongmei Han}, \bibinfo{person}{Fan Jiang}, \bibinfo{person}{Fei Wu}, {and} \bibinfo{person}{Yunliang Jiang}.} \bibinfo{year}{2023}\natexlab{}.
\newblock \showarticletitle{Examining the relationship between peer feedback classified by deep learning and online learning burnout}.
\newblock \bibinfo{journal}{\emph{Computers \& Education}}  \bibinfo{volume}{207} (\bibinfo{year}{2023}), \bibinfo{pages}{104910}.
\newblock
\href{https://doi.org/10.1016/j.compedu.2023.104910}{doi:\nolinkurl{10.1016/j.compedu.2023.104910}}


\bibitem[Huisman et~al\mbox{.}(2018)]%
        {huisman2018}
\bibfield{author}{\bibinfo{person}{Bart Huisman}, \bibinfo{person}{Nadira Saab}, \bibinfo{person}{Paul~Van den Broek}, {and} \bibinfo{person}{Jan~Van Driel}.} \bibinfo{year}{2018}\natexlab{}.
\newblock \showarticletitle{The impact of formative peer feedback on higher education students’ academic writing: a Meta-Analysis}.
\newblock \bibinfo{journal}{\emph{Assessment \& Evaluation in Higher Education}} \bibinfo{volume}{44}, \bibinfo{number}{6} (\bibinfo{year}{2018}), \bibinfo{pages}{863--880}.
\newblock
\href{https://doi.org/10.1080/02602938.2018.1545896}{doi:\nolinkurl{10.1080/02602938.2018.1545896}}


\bibitem[Koonchanok et~al\mbox{.}(2024)]%
        {koonchanok2024}
\bibfield{author}{\bibinfo{person}{Ratanond Koonchanok}, \bibinfo{person}{Yanling Pan}, {and} \bibinfo{person}{Hyeju Jang}.} \bibinfo{year}{2024}\natexlab{}.
\newblock \showarticletitle{Public attitudes toward chatgpt on twitter: sentiments, topics, and occupations}.
\newblock \bibinfo{journal}{\emph{Social Network Analysis and Mining}} \bibinfo{volume}{14}, \bibinfo{number}{1} (\bibinfo{year}{2024}), \bibinfo{pages}{106}.
\newblock
\showISSN{1869-5469}
\href{https://doi.org/10.1007/s13278-024-01260-7}{doi:\nolinkurl{10.1007/s13278-024-01260-7}}


\bibitem[Kreutel(2007)]%
        {kreutel2007}
\bibfield{author}{\bibinfo{person}{Karen Kreutel}.} \bibinfo{year}{2007}\natexlab{}.
\newblock \showarticletitle{"I’m not agree with you." ESL Learners’ Expressions of Disagreement}.
\newblock \bibinfo{journal}{\emph{Teaching English as a Second or Foreign Language}} \bibinfo{volume}{11}, \bibinfo{number}{3} (\bibinfo{year}{2007}), \bibinfo{pages}{1--35}.
\newblock


\bibitem[Lane et~al\mbox{.}(2018)]%
        {lane2018}
\bibfield{author}{\bibinfo{person}{Jacqueline~Ng Lane}, \bibinfo{person}{Bruce Ankenman}, {and} \bibinfo{person}{Seyed Iravani}.} \bibinfo{year}{2018}\natexlab{}.
\newblock \showarticletitle{Insight into Gender Differences in Higher Education: Evidence from Peer Reviews in an Introductory STEM Course}.
\newblock \bibinfo{journal}{\emph{Service Science}} \bibinfo{volume}{10}, \bibinfo{number}{4} (\bibinfo{year}{2018}), \bibinfo{pages}{442--456}.
\newblock
\href{https://doi.org/10.1287/serv.2018.0224}{doi:\nolinkurl{10.1287/serv.2018.0224}}


\bibitem[Leaper and Ayres(2007)]%
        {leaper2007}
\bibfield{author}{\bibinfo{person}{Campbell Leaper} {and} \bibinfo{person}{Melanie~M. Ayres}.} \bibinfo{year}{2007}\natexlab{}.
\newblock \showarticletitle{A Meta-Analytic Review of Gender Variations in Adults}.
\newblock \bibinfo{journal}{\emph{Personality and Social Psychology Review}} \bibinfo{volume}{11}, \bibinfo{number}{4} (\bibinfo{year}{2007}), \bibinfo{pages}{328--363}.
\newblock
\href{https://doi.org/10.1177/1088868307302221}{doi:\nolinkurl{10.1177/1088868307302221}}


\bibitem[McGarrell(2010)]%
        {mcgarrell2010}
\bibfield{author}{\bibinfo{person}{Hedy McGarrell}.} \bibinfo{year}{2010}\natexlab{}.
\newblock \showarticletitle{Native and non-native English speaking student teachers engage in peer feedback}.
\newblock \bibinfo{journal}{\emph{Canadian Journal of Applied Linguistics}} \bibinfo{volume}{13}, \bibinfo{number}{1} (\bibinfo{year}{2010}), \bibinfo{pages}{71--90}.
\newblock


\bibitem[McKenzie et~al\mbox{.}(2023)]%
        {mckenzie2023}
\bibfield{author}{\bibinfo{person}{Brian McKenzie}, \bibinfo{person}{Enyu Zhou}, {and} \bibinfo{person}{Alessandro Regio}.} \bibinfo{year}{2023}\natexlab{}.
\newblock \bibinfo{title}{Graduate Enrollment and Degrees: 2012 to 2022 Report}.
\newblock \bibinfo{howpublished}{\url{https://cgsnet.org/wp-content/uploads/2023/10/2022-Graduate-Enrollment-and-Degrees-Final-Report.pdf}}.
\newblock
\newblock
\shownote{Retrieved January 15, 2025}.


\bibitem[Neumann and Linzmayer(2021)]%
        {neumann2021}
\bibfield{author}{\bibinfo{person}{Marion Neumann} {and} \bibinfo{person}{Robin Linzmayer}.} \bibinfo{year}{2021}\natexlab{}.
\newblock \showarticletitle{Capturing Student Feedback and Emotions in Large Computing Courses: A Sentiment Analysis Approach}. In \bibinfo{booktitle}{\emph{Proceedings of the 52nd ACM Technical Symposium on Computer Science Education}} (Virtual Event, USA) \emph{(\bibinfo{series}{SIGCSE '21})}. \bibinfo{publisher}{Association for Computing Machinery}, \bibinfo{address}{New York, NY, USA}, \bibinfo{pages}{541–547}.
\newblock
\showISBNx{9781450380621}
\href{https://doi.org/10.1145/3408877.3432403}{doi:\nolinkurl{10.1145/3408877.3432403}}


\bibitem[Newman and Joyner(2018)]%
        {newman2018}
\bibfield{author}{\bibinfo{person}{Heather Newman} {and} \bibinfo{person}{David Joyner}.} \bibinfo{year}{2018}\natexlab{}.
\newblock \showarticletitle{Sentiment Analysis of Student Evaluations of Teaching}. In \bibinfo{booktitle}{\emph{Artificial Intelligence in Education. AIED 2018. Lecture Notes in Computer Science, vol 10948}}, \bibfield{editor}{\bibinfo{person}{C.~Penstein Rosé}, \bibinfo{person}{R.~Martínez-Maldonado}, \bibinfo{person}{H.~Hoppe}, \bibinfo{person}{R.~Luckin}, \bibinfo{person}{M.~Mavrikis}, \bibinfo{person}{K.~Porayska-Pomsta}, \bibinfo{person}{B.~McLaren}, {and} \bibinfo{person}{B.~du~Boulay}} (Eds.). \bibinfo{publisher}{Springer, Cham}, \bibinfo{address}{Cham, Switzerland}, \bibinfo{pages}{246--250}.
\newblock
\href{https://doi.org/10.1007/978-3-319-93846-2_45}{doi:\nolinkurl{10.1007/978-3-319-93846-2_45}}


\bibitem[NFAP(2021)]%
        {nfap2021}
\bibfield{author}{\bibinfo{person}{NFAP}.} \bibinfo{year}{2021}\natexlab{}.
\newblock \bibinfo{title}{International Students in Science and Engineering}.
\newblock \bibinfo{howpublished}{\url{https://nfap.com/wp-content/uploads/2021/09/International-Students-in-Science-and-Engineering.NFAP-Policy-Brief.August-2021.pdf}}.
\newblock
\newblock
\shownote{Retrieved January 15, 2025}.


\bibitem[Nielsen(2014)]%
        {nielsen2014}
\bibfield{author}{\bibinfo{person}{Kathryn Nielsen}.} \bibinfo{year}{2014}\natexlab{}.
\newblock \showarticletitle{On class, race, and dynamics of privilege: Supporting generation 1.5 writers across the curriculum}.
\newblock In \bibinfo{booktitle}{\emph{WAC and second language writers: Research towards linguistically and culturally inclusive programs and practices}}. \bibinfo{publisher}{Parlor Press}, \bibinfo{address}{Anderson, SC}, \bibinfo{pages}{129--150}.
\newblock


\bibitem[Noroozi et~al\mbox{.}(2024)]%
        {noroozi2024}
\bibfield{author}{\bibinfo{person}{Omid Noroozi}, \bibinfo{person}{Maryam Alqassab}, \bibinfo{person}{Nafiseh~Taghizadeh Kerman}, \bibinfo{person}{Seyyed~Kazem Banihashem}, {and} \bibinfo{person}{Ernesto Panadero}.} \bibinfo{year}{2024}\natexlab{}.
\newblock \showarticletitle{Does perception mean learning? Insights from an online peer feedback setting}.
\newblock \bibinfo{journal}{\emph{Assessment \& Evaluation in Higher Education}} \bibinfo{volume}{50}, \bibinfo{number}{1} (\bibinfo{year}{2024}), \bibinfo{pages}{83--97}.
\newblock
\href{https://doi.org/10.1080/02602938.2024.2345669}{doi:\nolinkurl{10.1080/02602938.2024.2345669}}


\bibitem[Ocampo et~al\mbox{.}(2023)]%
        {ocampo2023}
\bibfield{author}{\bibinfo{person}{José~Carlos Ocampo}, \bibinfo{person}{Ernesto Panadero}, \bibinfo{person}{David Zamorano}, \bibinfo{person}{Iván Sánchez-Iglesias}, {and} \bibinfo{person}{Fernando~Diez Ruiz}.} \bibinfo{year}{2023}\natexlab{}.
\newblock \showarticletitle{The effects of gender and training on peer feedback characteristics}.
\newblock \bibinfo{journal}{\emph{Assessment \& Evaluation in Higher Education}} \bibinfo{volume}{49}, \bibinfo{number}{4} (\bibinfo{year}{2023}), \bibinfo{pages}{539--555}.
\newblock
\href{https://doi.org/10.1080/02602938.2024.2345669}{doi:\nolinkurl{10.1080/02602938.2024.2345669}}


\bibitem[Panadero and Lipnevich(2022)]%
        {panadero2022}
\bibfield{author}{\bibinfo{person}{Ernesto Panadero} {and} \bibinfo{person}{Anastasiya~A. Lipnevich}.} \bibinfo{year}{2022}\natexlab{}.
\newblock \showarticletitle{A review of feedback models and typologies: Towards an integrative model of feedback elements}.
\newblock \bibinfo{journal}{\emph{Educational Research Review}}  \bibinfo{volume}{35} (\bibinfo{year}{2022}), \bibinfo{pages}{100416}.
\newblock
\href{https://doi.org/10.1016/j.edurev.2021.100416}{doi:\nolinkurl{10.1016/j.edurev.2021.100416}}


\bibitem[Reichelt(2021)]%
        {reichelt2021}
\bibfield{author}{\bibinfo{person}{Melinda Reichelt}.} \bibinfo{year}{2021}\natexlab{}.
\newblock \showarticletitle{Linguistic Bias against ESL Writing?}
\newblock In \bibinfo{booktitle}{\emph{Linguistic Discrimination in US Higher Education}}. \bibinfo{publisher}{Routledge}, \bibinfo{address}{New York, NY}, \bibinfo{pages}{20--37}.
\newblock


\bibitem[Rubin and Williams-James(1997)]%
        {rubin1997}
\bibfield{author}{\bibinfo{person}{Donald~L. Rubin} {and} \bibinfo{person}{Melanie Williams-James}.} \bibinfo{year}{1997}\natexlab{}.
\newblock \showarticletitle{The impact of writer nationality on mainstream teachers}.
\newblock \bibinfo{journal}{\emph{Journal of Second Language Writing}} \bibinfo{volume}{6}, \bibinfo{number}{2} (\bibinfo{year}{1997}), \bibinfo{pages}{139--154}.
\newblock
\href{https://doi.org/10.1016/s1060-3743(97)90031-x}{doi:\nolinkurl{10.1016/s1060-3743(97)90031-x}}


\bibitem[Wärnsby et~al\mbox{.}(2018)]%
        {warnsby2018}
\bibfield{author}{\bibinfo{person}{Anna Wärnsby}, \bibinfo{person}{Asko Kauppinen}, \bibinfo{person}{Laura Aull}, \bibinfo{person}{Djuddah Leijen}, {and} \bibinfo{person}{Joe Moxley}.} \bibinfo{year}{2018}\natexlab{}.
\newblock \showarticletitle{Affective Language in Student Peer Reviews: Exploring Data from Three Institutional Contexts}.
\newblock \bibinfo{journal}{\emph{Journal of Academic Writing}} \bibinfo{volume}{8}, \bibinfo{number}{1} (\bibinfo{year}{2018}), \bibinfo{pages}{28--53}.
\newblock
\href{https://doi.org/10.18552/joaw.v8i1.429}{doi:\nolinkurl{10.18552/joaw.v8i1.429}}


\bibitem[Zhou et~al\mbox{.}(2019)]%
        {zhou2019}
\bibfield{author}{\bibinfo{person}{Jiming Zhou}, \bibinfo{person}{Yongyan Zheng}, {and} \bibinfo{person}{Joanna Hong-Meng Tai}.} \bibinfo{year}{2019}\natexlab{}.
\newblock \showarticletitle{Grudges and gratitude: the social-affective impacts of peer assessment}.
\newblock \bibinfo{journal}{\emph{Assessment \& Evaluation in Higher Education}} \bibinfo{volume}{45}, \bibinfo{number}{3} (\bibinfo{year}{2019}), \bibinfo{pages}{345--358}.
\newblock
\href{https://doi.org/10.1080/02602938.2019.1643449}{doi:\nolinkurl{10.1080/02602938.2019.1643449}}


\bibitem[Zhu(2001)]%
        {zhu2001}
\bibfield{author}{\bibinfo{person}{Wei Zhu}.} \bibinfo{year}{2001}\natexlab{}.
\newblock \showarticletitle{Interaction and feedback in mixed peer response groups}.
\newblock \bibinfo{journal}{\emph{Journal of Second Language Writing}} \bibinfo{volume}{10}, \bibinfo{number}{4} (\bibinfo{year}{2001}), \bibinfo{pages}{251--276}.
\newblock
\href{https://doi.org/10.1016/s1060-3743(01)00043-1}{doi:\nolinkurl{10.1016/s1060-3743(01)00043-1}}


\end{thebibliography}
\end{document}